%% file: paper1780.tex
\documentclass[runningheads]{llncs}
\usepackage{graphicx}
\usepackage{import}
\usepackage{amsmath}
\usepackage{upgreek}
\usepackage[dvipsnames,table,x11names]{xcolor}

\usepackage{tabularx,booktabs}
\usepackage{arydshln}
\usepackage{tabu}
\usepackage{adjustbox}
\usepackage{enumerate}
\usepackage{multirow}
\usepackage{multicol} 
\usepackage{enumitem}
\usepackage{array}
\usepackage{gensymb}
\usepackage{float}
\usepackage{balance}
\usepackage{xspace}
\usepackage{amssymb}
\usepackage{pifont}
\usepackage{hyperref}
\usepackage{mathtools}
\usepackage{xfrac}
\usepackage{scalerel}
\hypersetup{
    colorlinks,
    linkcolor={red!50!black},
    citecolor={blue!50!black},
    urlcolor={black}
}

\newcolumntype{P}[1]{>{\centering\arraybackslash}p{#1}}

\makeatletter
\newcommand\avsuminner[2]{%
  {\sbox0{$\m@th#1\sum$}%
   \vphantom{\usebox0}%
   \ooalign{%
     \hidewidth
     \smash{\vrule height\dimexpr\ht0+1pt\relax depth\dimexpr\dp0+1pt\relax}%
     \hidewidth\cr
     $\m@th#1\sum$\cr
   }%
  }%
}
\makeatother
\usepackage{soul}
\usepackage{url}
\usepackage[]{hyperref}
\usepackage[utf8]{inputenc}
\usepackage{booktabs}
\usepackage{pgfplots}
\usepackage{pgfplotstable}
\usetikzlibrary{pgfplots.groupplots}
\usepgfplotslibrary{colorbrewer}
\pgfplotsset{/pgfplots/error bars/error bar style={black,thick}}
\definecolor{shadecolor}{rgb}{0.95,0.95,0.95}

\pgfplotsset{compat=1.11,
        /pgfplots/ybar legend/.style={
        /pgfplots/legend image code/.code={%
        \draw[##1,/tikz/.cd,bar width=3pt,yshift=-0.2em,bar shift=0pt]
                plot coordinates {(0cm,0.8em)};},
},}

\input{defs}

\renewcommand{\orcidID}[1]{\href{https://orcid.org/#1}{\includegraphics[scale=0.0075]{img/orcid.jpg}}}
\usepackage{todonotes}
\usepackage{arydshln} 

\usepackage[T1]{fontenc}

\graphicspath{{figures/}}
\usepackage{stackengine}
\newcommand\textsub[1]{\stackengine{-.5ex}{}{\scriptsize#1}{O}{l}{F}{F}{L}}
\DeclareMathOperator{\argmaxH}{argmax}  
\DeclareMathOperator{\maxH}{max}  
\begin{document}
\title{Domain Adaptation for Medical Image Segmentation using Transformation-Invariant Self-Training\thanks{This work was funded by Haag-Streit Switzerland.}}
\titlerunning{Transformation-Invariant Self-Training}
%
\author{Negin Ghamsarian\inst{1} 
\and Javier Gamazo Tejero\inst{1} 
\and Pablo Márquez Neila \inst{1} 
\and Sebastian Wolf \inst{2} 
\and Martin Zinkernagel \inst{2} 
\and Klaus Schoeffmann \inst{3} 
\and Raphael Sznitman \inst{1} 
} 

\institute{Center for AI in Medicine, Faculty of Medicine, University of Bern, Switzerland\\ 
\and Department of Ophthalmology, Inselspital, Bern, Switzerland\\
\and Department of Information Technology, Klagenfurt University, Austria\\
\email{negin.ghamsarian@unibe.ch}
}

\authorrunning{N. Ghamsarian et al.}
\maketitle           

\begin{abstract}
Models capable of leveraging unlabelled data are crucial in overcoming large distribution gaps between the acquired datasets across different imaging devices and configurations. In this regard, self-training techniques based on pseudo-labeling have been shown to be highly effective for semi-supervised domain adaptation. However, the unreliability of pseudo labels can hinder the capability of self-training techniques to induce abstract representation from the unlabeled target dataset, especially in the case of large distribution gaps. 
Since the neural network performance should be invariant to image transformations, we look to this fact to identify uncertain pseudo labels. Indeed, we argue that transformation invariant detections can provide more reasonable approximations of ground truth. Accordingly, we propose a semi-supervised learning strategy for domain adaptation termed transformation-invariant self-training (TI-ST). The proposed method assesses pixel-wise pseudo-labels' reliability and filters out unreliable detections during self-training. We perform comprehensive evaluations for domain adaptation using three different modalities of medical images, two different network architectures, and several alternative state-of-the-art domain adaptation methods. Experimental results confirm the superiority of our proposed method in mitigating the lack of target domain annotation and boosting segmentation performance in the target domain.

\keywords{Semi-Supervised Learning . Domain Adaptation  \and Semantic Segmentation \and Self Training . Cataract Surgery . MRI . OCT .}
\end{abstract}

\input{01_introduction}

\input{02_methodology}

\input{03_experimental_setup}

\input{04_experimental_results}

\input{05_conclusion}

\bibliographystyle{splncs04}
\bibliography{bib}
\end{document}

%% file: defs.tex
\newif\ifdraft
\drafttrue

\definecolor{orange}{rgb}{1,0.5,0}
\definecolor{pink}{rgb}{0.98, 0.38, 0.5}
\definecolor{darkgreen}{rgb}{0.055, 0.490, 0.016} 

\ifdraft
 \usepackage[normalem]{ulem}
 \newcommand{\RS}[1]{{\color{red}{\bf RS: #1}}}
 
 \newcommand{\PMN}[1]{{\color{orange}{\bf PMN: #1}}}
 
 \newcommand{\JGT}[1]{{\color{blue} JGT: #1}}

\else
 \newcommand{\sout}[1]{}
 \newcommand{\RS}[1]{{\color{red}{}}}
 
 \newcommand{\PMN}[1]{{\color{red}{}}}
 
 \newcommand{\JGT}[1]{{\color{blue}{}}}
 
\fi

\newcolumntype{S}{>{\centering\arraybackslash}p{1.05cm}}
\newcolumntype{L}{>{\small}l}
\newcolumntype{T}{>{\centering\arraybackslash}p{1.1cm}}

\newcommand{\comment}[1]{}

%% file: 01_introduction.tex
\section{Introduction}
\label{sec:introduction}

Semantic segmentation is a prerequisite for a broad range of medical imaging applications, including disease diagnosis and treatment~\cite{CT-free}, surgical workflow analysis~\cite{LocalPhase,RBE}, operation room planning, and surgical outcome prediction~\cite{LensID}. While supervised deep learning approaches have yielded satisfactory performance in semantic segmentation ~\cite{ghamsarian2022deeppyramid,ghamsarian2021recal}, their performance is heavily limited by the labeled training dataset distribution. Indeed, a network trained on a dataset acquired with a specific device or configuration can dramatically underperform when evaluated on a different device or conditions. Overcoming this entails new annotations per device, a demand that is hard to meet, especially for semantic segmentation, and even more so in the medical domain, where expert knowledge is essential.

Driven by the need to overcome this challenge, numerous semi-supervised learning paradigms have looked to alleviate annotation requirements in the target domain. Semi-supervised learning refers to methods that encourage learning abstract representations from an unlabeled dataset and extending the decision boundaries towards a more-generalized or target dataset distribution. These techniques can be categorized into (i) consistency regularization~\cite{CPS,TESSL,HCS,TCSM-V2,UDAMIS,TTUDA}, (ii) contrastive learning~\cite{CL-GLF,UDA-OCT}, (iii) adversarial learning~\cite{TTUDA}, and (iv) self-training \cite{st++,Reciprocal,UDACB}. Consistency regularization techniques aim to inject knowledge via penalizing inconsistencies for identical images that have undergone different distortions, such as transformations or dropouts, or fed into networks with different initializations~\cite{CPS}. Specifically, the $\Uppi$ model~\cite{TESSL} penalizes differences between the predictions of two transformed versions of each input image to reinforce consistent and augmentation-invariant predictions. Temporal ensembling~\cite{TESSL} is designed to alleviate the negative effect of noisy predictions by integrating predictions of consecutive training iterations. Cross-pseudo supervision regularizes the networks by enforcing similar predictions from differently initialized networks. 

More recent deep self-training approaches based on pseudo labels have emerged as promising techniques for unsupervised domain adaptation. These techniques assume that a trained network can approximate the ground-truth labels for unlabeled images. Since no metric guarantees pseudo-label reliability, several methods have been developed to alleviate pseudo-label error back-propagation. To progressively improve pseudo-labeling performance, reciprocal learning~\cite{Reciprocal} adopts a teacher-student framework where the student network performance on the source domain drives the teacher network weights updates. ST++~\cite{st++} proposes to evaluate the reliability of image-based pseudo labels based on the consistency of predictions in different network checkpoints. Subsequently, half of the more reliable images are utilized to re-train the network in the first step, and the trained network is used for pseudo-labeling the whole dataset for a second re-training step. Despite the effectiveness of state-of-the-art pseudo-labeling strategies, we argue that one important aspect has been underexplored: how can a trained network self-assess the reliability of its pixel-level predictions?

To this end, we propose a novel self-training framework with a self-assessment strategy for pseudo-label reliability. The proposed framework uses transformation-invariant highly-confident predictions in the target dataset for self-training. This objective is achieved by considering an ensemble of high-confidence predictions from transformed versions of identical inputs. To validate the effectiveness of our proposed framework on a variety of tasks, we evaluate our approach on three different semantic segmentation imaging modalities, including video (cataract surgery), optical coherence tomography (retina), and MRI (prostate). We perform comprehensive experiments to validate the performance of the proposed framework, namely ``Transformation-Invariant Self-Training''\footnote{The PyTorch implementation of TI-ST is publicly available at \url{https://github.com/Negin-Ghamsarian/Transformation-Invariant-Self-Training-MICCAI23}.} (TI-ST). The experimental results indicate that TI-ST significantly improves segmentation performance for unlabeled target datasets compared to numerous state-of-the-art alternatives.
\begin{figure}[t]
\centering
\includegraphics[width=0.95\textwidth]{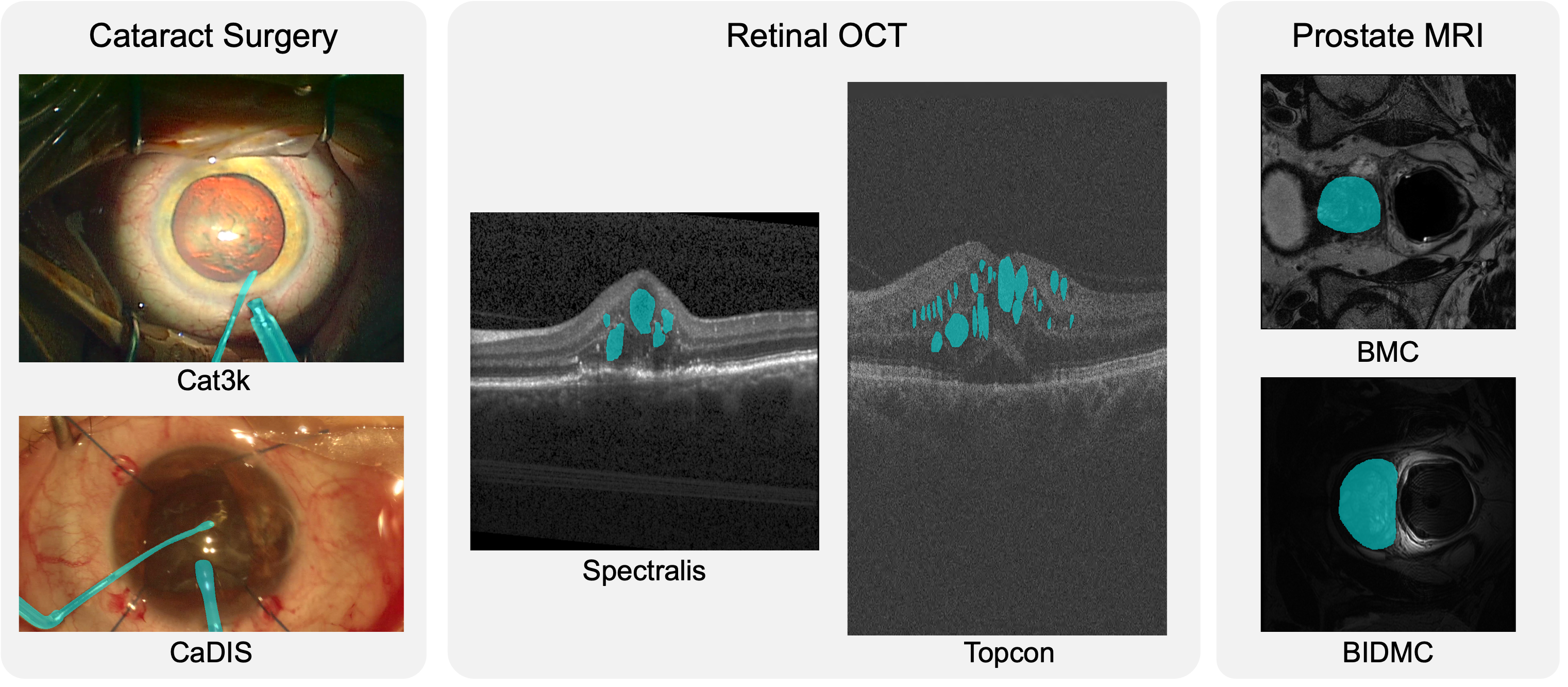}
\caption{Example images from the three adopted datasets: (1) cross-device-and-center instrument segmentation in cataract surgery videos (Cat101 vs. CaDIS), cross-device fluid segmentation in OCT (Spectralis vs. Topcon), and cross-institution prostate segmentation in MRI (BMC vs. BIDMC).
}
\label{fig:datasets}
\end{figure}

%% file: 02_methodology.tex
\section{Methodology}
\label{sec: Methodology}

Consider a labeled source dataset, $\mathcal{S}$, with training images $\mathcal{X_S}$ and corresponding segmentation labels $\mathcal{Y_S}$, while we denote a target dataset $\mathcal{T}$, containing only target images $\mathcal{X_T}$. We aim to train a network using $\mathcal{X_S}$, $\mathcal{Y_S}$, and $\mathcal{X_T}$ for semantic segmentation in the target dataset.

\begin{figure}[t]
\centering
\includegraphics[width=\textwidth]{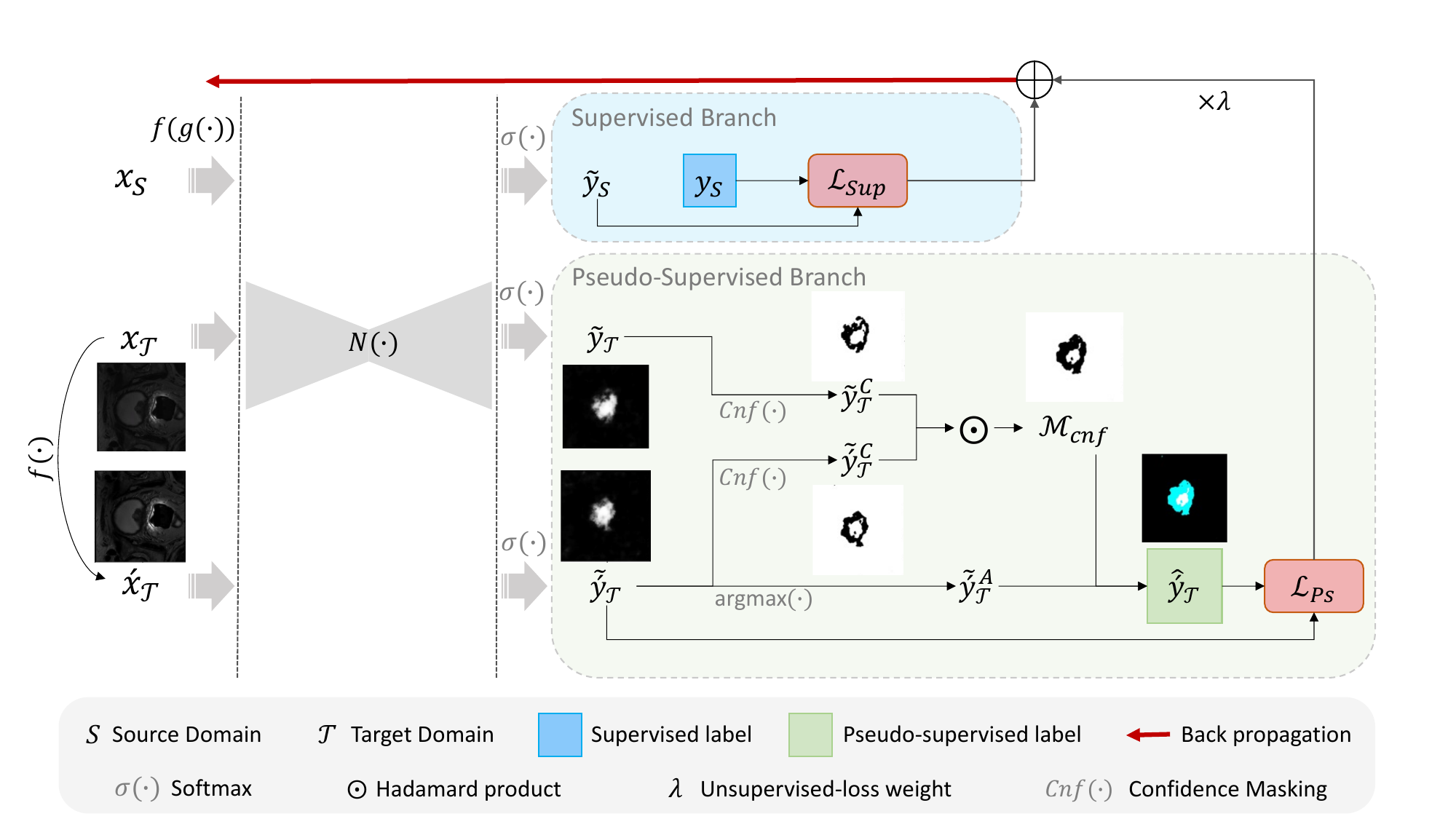}
\caption{Overview of the proposed semi-supervised domain adaptation framework based on transformation-invariant self-training (TI-ST). Ignored pseudo-labels during unsupervised loss computation are shown in turquoise.
}
\label{fig:BD}
\end{figure}

We propose to train the model using a self-supervised approach on the images $\mathcal{X_T}$ by assigning pseudo labels during training. Typical pseudo labels are computed from independent predictions of unlabeled images. Instead, our proposed framework adopts a self-assessment strategy to determine the reliability of predictions in an unsupervised fashion. Specifically, we propose to target highly-reliable predictions generated by a network aiming for transformation-invariant confidence. Compared to self-ensembling strategies that penalize the distant predictions corresponding to the transformed versions of identical inputs, our goal is to filter out transformation-variant predictions. Indeed, our method reinforces the ensemble of high-confidence predictions from two versions of the same target sample. Our proposed TI-ST framework simultaneously trains on the source and target domains, so as to progressively bridge the intra-domain distribution gap. Fig.~\ref{fig:BD} depicts our TI-ST framework, which we detail in the following sections. 

\subsection{Model} At training time, images from the source dataset are augmented using spatial $g(\cdot)$ and non-spatial $f(\cdot)$ transformations and passed through a segmentation network, $N(\cdot)$, by which the network is trained using a standard supervision loss. At the same time, images from the target dataset are also passed to the network. Specifically, we feed two versions of each target image to the network: (1) the original target image $x_\mathcal{T}$, and (2) its non-spatially transformed version, $\Acute{x_\mathcal{T}} = f(x_\mathcal{T})$. 
Once fed through the network, the corresponding predictions can be defined as $\tilde{y_{\mathcal{T}}} = \sigma(N(x_{\mathcal{T}}))$ and $\tilde{\acute{y_{\mathcal{T}}}} = \sigma(N(\Acute{x_{\mathcal{T}}}))$, where $\sigma(\cdot)$ is the Softmax operation. We then define a confidence-mask ensemble as
\begin{equation}
\mathcal{M}_{cnf} = 
Cnf(\tilde{{y_{\mathcal{T}}}})
\odot
Cnf(\tilde{\Acute{y_{\mathcal{T}}}}),
\label{eq: ensemble of confidence}
\end{equation}
\noindent
where $\odot$ refers to Hadamard product used for element-wise multiplication, and $Cnf$ is the high confidence masking function,
\begin{equation}
    Cnf_{\textsub{$\in (W\times H$)}}(y) =
    \begin{cases}
    1, & \text{if    } \maxH_{\textsub{C}}(y) > \uptau\\
    0, & \text{else.  }
    \end{cases}
    \label{eq: filtering}
\end{equation}
\noindent
where $\uptau \in (0.5,1) $ is the confidence threshold, and $H$, $W$, and $C$ are the height, width, and number of classes in the output, respectively. Specifically, $\mathcal{M}_{cnf}$ encodes regions of confident predictions that are invariant to transformations.
We can then compute the pseudo-ground-truth mask for each input from the target dataset as
\begin{equation}
\hat{\Acute{y_{\mathcal{T}}}} = 
\begin{cases}
\argmaxH_{\textsub{C}} (\tilde{\Acute{y_{\mathcal{T}}}}), & \text{if  } \:\mathcal{M}_{cnf} = 1\\
\text{ignore}, & \text{else.  }
\end{cases}
\end{equation}
\noindent

\subsection{Training}
To train our model, we simultaneously consider both the source and target samples by minimizing the following loss,
\begin{equation}
    \mathcal{L}_{overall} = \mathcal{L}_{Sup}( \tilde{y_\mathcal{S}}, y_\mathcal{S}) + \lambda \Big(\mathcal{L}_{Ps}(\tilde{\Acute{y_{\mathcal{T}}}}, \hat{\Acute{y_{\mathcal{T}}}})\Big) ,
    \label{eq: loss}
\end{equation}
\noindent
where $\mathcal{L}_{Sup}$ and $\mathcal{L}_{Ps}$ indicate the supervised and pseudo-supervised loss functions used, respectively. We set $\lambda$ as a time-dependent weighing function that gradually increases the share of pseudo-supervised loss. Intuitively, our pseudo-supervised loss enforces predictions on transformation-invariant highly-confident regions for unlabeled images. 

\subsubsection{Discussion:} 
The quantity and distribution of supervised data are determining factors in neural networks' performance. With highly distributed large-scale supervisory data, neural networks converge to an optimal state efficiently. However, when only limited supervisory data with heterogeneous distribution from the inference dataset are available, using more sophisticated methods to leverage a priori knowledge is essential. Our proposed use of invariance of network predictions with respect to data augmentation is a strong form of knowledge that can be learned through dataset-dependent augmentations. The trained network is then expected to provide consistent predictions under diverse transformations. Hence, the transformation variance of the network predictions can indicate the network's prediction doubt and low confidence correspondingly. We take advantage of this characteristic to assess the reliability of predictions and filter out unreliable pseudo-labels.

%% file: 03_experimental_setup.tex
\section{Experimental setup}
\label{sec: experimental settings}

\subsubsection{Datasets:} We validate our approach on three cross-device/site datasets for three different modalities: 

\begin{itemize}
    \item \textbf{Cataract:} instrument segmentation in cataract surgery videos~\cite{CaDIS,cat101}. We set the ``Cat101''~\cite{cat101} as the source dataset and the ``CaDIS'' as the target domain dataset \cite{CaDIS}. 
    \item \textbf{OCT: }IRF Fluid segmentation in retinal OCTs~\cite{RETOUCH}. We use the high-quality ``Spectralis'' dataset as the source and the lower-quality ``Topcon'' dataset as the target domain.
    \item \textbf{MRI:} multi-site prostate segmentation~\cite{s-net}. We sample volumes from ``BMC'' and ``BIDMC'' as the source and target domain, respectively. 
\end{itemize} 

We follow a four-fold validation strategy for all three cases and report the average results over all folds. The average number of labeled training images (from the source domain), unlabeled training images (from the target domain), and test images per fold are equal to ($207, 3189,58$) for Cataract, ($391,569,115$) for OCT, and ($273,195,65$) for MRI dataset.

\subsubsection{Baseline methods:} We compare the performance of our proposed transformation-invariant self-training (SI-ST) method against seven state-of-the-art semi-supervised learning methods: $\Uppi$ models~\cite{TESSL}, temporal ensembling~\cite{TESSL}, mean teacher~\cite{SE}, cross pseudo supervision (CSP)~\cite{CPS}, reciprocal learning (RL)~\cite{Reciprocal}, self-training (ST)~\cite{st++}, and mutual correction framework (MCF)~\cite{wang2023mcf}.

\subsubsection{Networks and training settings:} We evaluate our TI-ST framework using two different architectures:  (1) DeepLabV3+~\cite{DeepLabV3} with ResNet50 backbone~\cite{ResNet} and (2) scSE~\cite{scSE} with VGG16 backbone. Both backbones are initialized with the ImageNet~\cite{ImageNet} pre-trained parameters. We use a batch size of four for the Cataract and MRI datasets and a batch size of two for the OCT dataset. For all training strategies, we set the number of epochs to 100. The initial learning rate is set to 0.001 and decayed by a factor of $\gamma = 0.8$ every two epochs. The input size of the networks is $512\times 512$ for cataract and OCT and $384\times 384$ for the MRI dataset. 
As spatial transformations $g(\cdot)$, we apply cropping and random rotation (up to 30 degrees). 
The non-spatial transformations, $f(\cdot)$, include color jittering (brightness = 0.7, contrast = 0.7, saturation = 0.7), Gaussian blurring, and random sharpening. The confidence threshold $\uptau$ for the self-training framework and the proposed TI-ST framework is set to $0.85$ except in the ablation studies (See the next section). In Eq.~\eqref{eq: loss}, the weighting function $\lambda$ ramps up from the first epoch along a Gaussian curve equal to $\exp[-5(1-\text{current-epoch}/{\text{total-epochs}})]$. The self-supervised loss is set to the cross-entropy loss, and the supervised loss is set to the \textit{cross entropy log dice} loss, which is a weighted sum of cross-entropy and the logarithm of soft dice coefficient. For the TI-ST framework, we only use non-spatial transformations for the self-training branch for simplicity.

%% file: 04_experimental_results.tex
\input{tables/All}
\begin{figure}[b]
\centering
\includegraphics[width=1\textwidth]{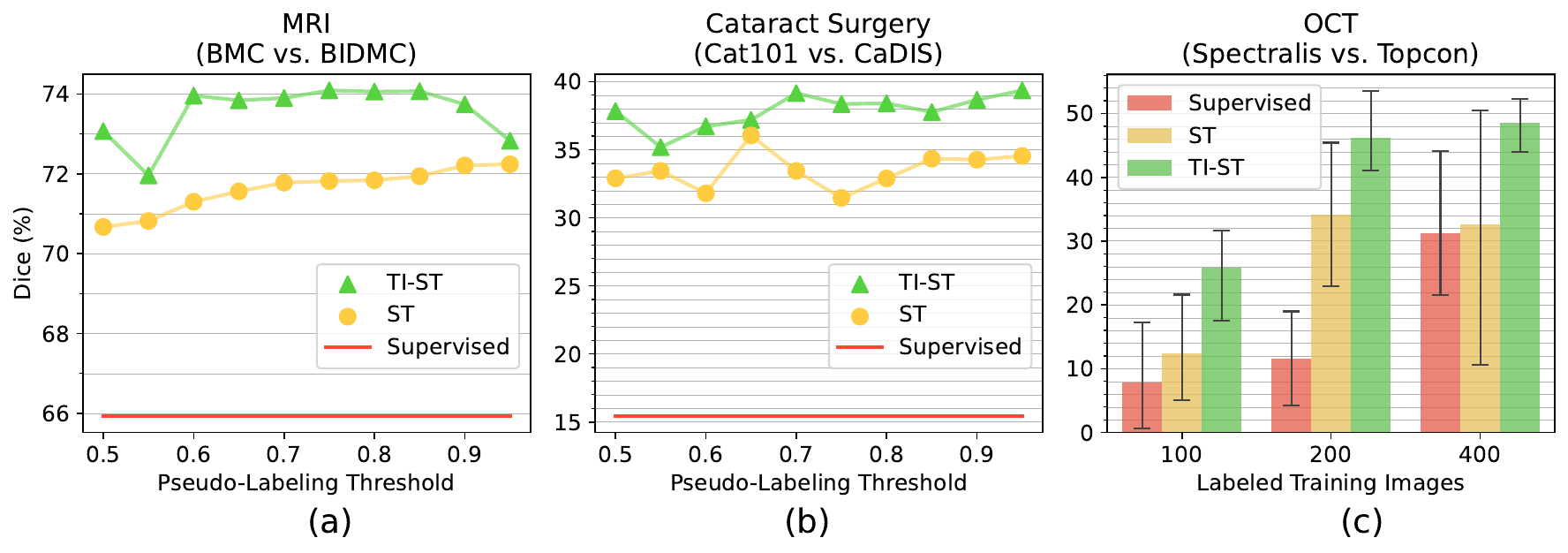}
\caption{Ablation studies on the pseudo-labeling threshold and size of the labeled dataset. 
}
\label{fig:ablation}
\end{figure}

\begin{figure}[t]
\centering
\includegraphics[width=1\textwidth]{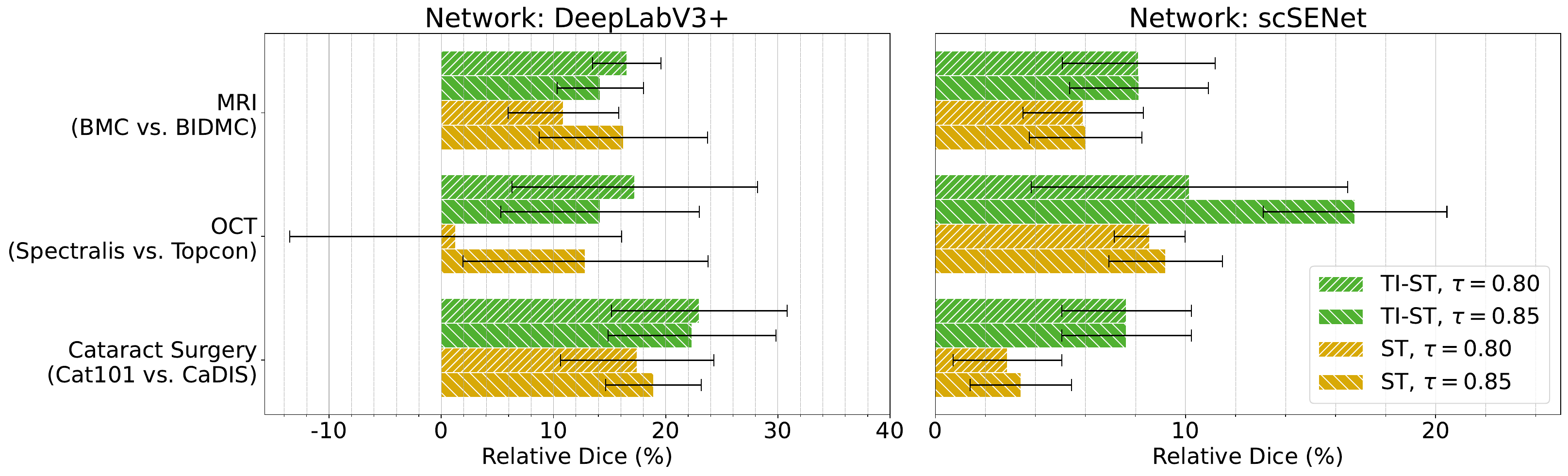}
\caption{Ablation study on the performance stability of TI-ST vs. ST across the different experimental segmentation tasks.
}
\label{fig:ablation_stability}
\end{figure}

\begin{figure}[t]
\centering
\includegraphics[width=.9\textwidth]{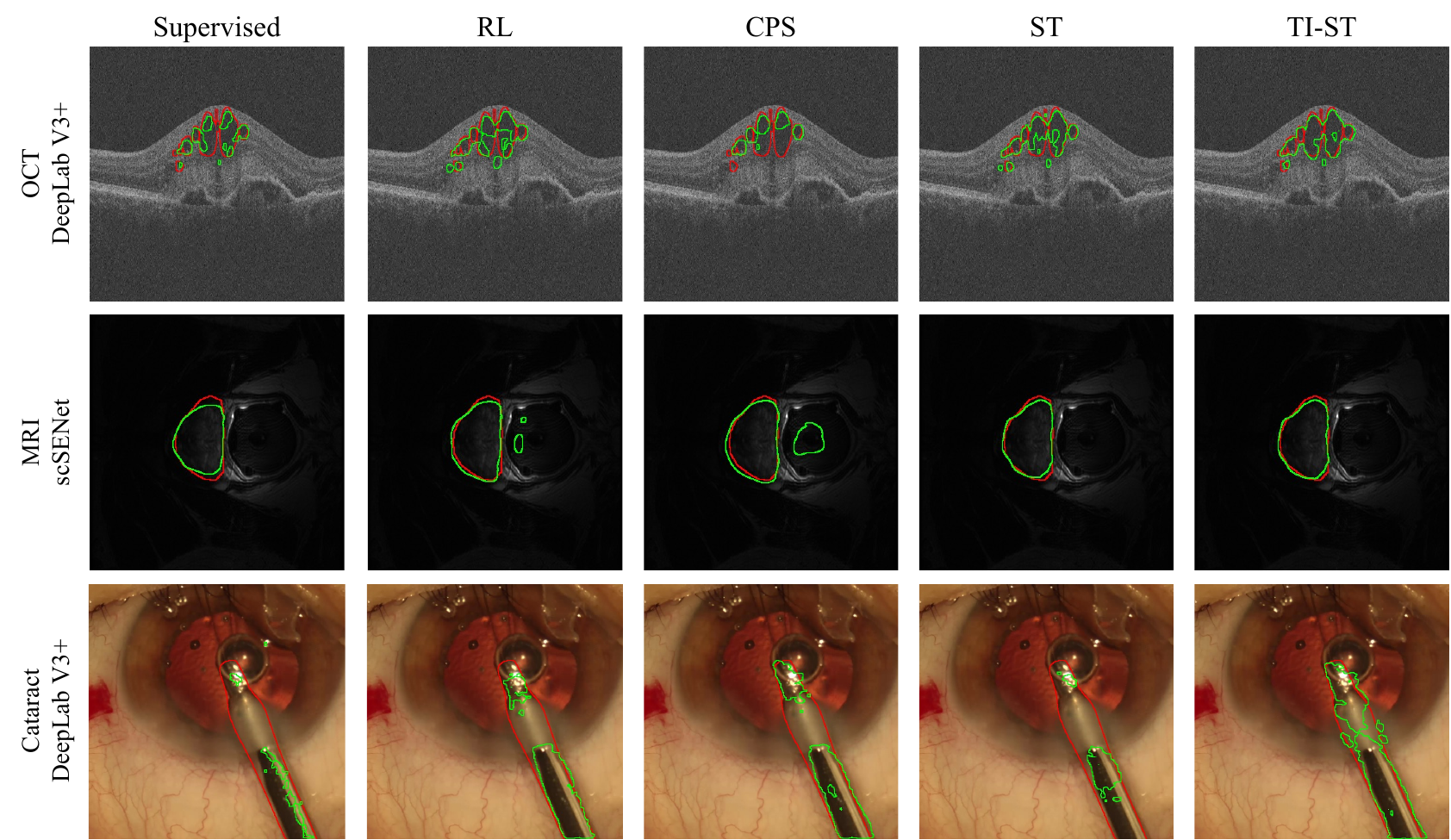}
\caption{Qualitative comparisons between the performance of TI-ST and four existing methods.
}
\label{fig:qualitative}
\end{figure}
\section{Results}
\label{sec: experimental results}

Table~\ref{tab:quantitative} compares the performance of our transformation-invariant self-training (TI-ST) approach with alternative methods across three tasks and using two network architectures. According to the quantitative results, TI-ST, RL, ST, and CPS are the best-performing methods. Nevertheless, our proposed TI-ST achieves the highest average relative improvement in dice score compared to naive supervised learning ($16.18\%$ average improvement). Considering our main competitor (RL), we note that our proposed TI-ST method is a one-stage framework using one network. In contrast, RL is a two-stage framework (requiring a pre-training stage) and uses a teacher-student network. Hence, TI-ST is also more efficient than RL in terms of time and computation.  Furthermore, the proposed strategy demonstrates the most consistent results when evaluated on different tasks, regardless of the utilized neural network architecture. 

Fig.~\ref{fig:ablation}-(a-b) demonstrates the effect of the pseudo-labeling threshold on TI-ST performance compared with regular ST. We observe that filtering out unreliable pseudo-labels based on transformation variance can remarkably boost pseudo-supervision performance regardless of the threshold. Fig.~\ref{fig:ablation}-(c) compares the performance of the supervised baseline, ST, and TI-ST with respect to the number of source-domain labeled training images. While ST performance converges when the number of labeled images increases, our TI-ST pushes decision boundaries toward the target domain dataset by avoiding training with transformation variant pseudo-labels. We validates the stability of TI-ST vs. ST  with different labeling thresholds (0.80 and 0.85) over four training folds in Fig.~\ref{fig:ablation_stability}, where TI-ST achieves a higher average improvement relative to supervised learning for different tasks and network architectures. This analysis also shows that the performance of ST is sensitive to the pseudo-labeling threshold and generally degrades by reducing the threshold due to resulting in wrong pseudo labels. However, TI-ST can effectively ignore false predictions in lower thresholds and take advantage of a higher amount of correct pseudo labels. This superior performance is depicted qualitatively in Fig.~\ref{fig:qualitative}.

%% file: tables/All.tex
\begin{table}[t]
\centering
\caption{Quantitative comparisons in Dice score (\%) among the proposed (TI-ST) and alternative methods for DeepLabV3+~\cite{DeepLabV3} (DLV3+) and scSENet~\cite{scSE} and the three datasets. Relative Dice computed over the Supervised baseline. The best results are shown in \textcolor{ForestGreen}{green}.}
\label{tab:quantitative}
\begin{tabular}{lm{1.3cm}*{7}{>{\centering\arraybackslash}m{1.3cm}}}
\toprule
Modality & \multicolumn{2}{c}{\footnotesize{Cataract Surgery}} & \multicolumn{2}{c}{\footnotesize{OCT}} & \multicolumn{2}{c}{\footnotesize{MRI}} & \multicolumn{1}{l}{\multirow{2}{*}{\footnotesize{Avg. Rel.}}} \\ \cmidrule(lr){2-3}\cmidrule(lr){4-5}\cmidrule(lr){6-7}
Network & \footnotesize{DLV3+} & \footnotesize{scSENet} & \footnotesize{DLV3+} & \footnotesize{scSENet} & \footnotesize{DLV3+} & \footnotesize{scSENet} &   \\ \midrule
Supervised & 15.42 & 37.67 & 22.87 & 24.08 & 52.39 & 65.93 & N/A \\
\rowcolor{shadecolor}$\Uppi$ Model~\cite{TESSL} & 27.55 & 35.56 & 1.12 & 0.00 & 10.00 & 6.87 & -22.88 \\
TE~\cite{TESSL} & 33.10 & 42.32 & 42.13 & 39.86 & 63.41 & 67.25 & 11.62 \\
\rowcolor{shadecolor}Mean Teacher~\cite{SE} & 11.06 & 39.54 & 19.11 & 4.70 & 64.82 & 66.87 & -2.04 \\
RL~\cite{Reciprocal} & 34.40 & 45.13 & 48.73 & \textcolor{ForestGreen}{47.70} & 60.79 & 70.20 & 14.77 \\
\rowcolor{shadecolor}CPS~\cite{CPS} & 36.24 & 39.40 & 47.31 & 14.71 & \textcolor{ForestGreen}{76.00} & 68.80 & 10.68 \\
ST~\cite{st++} & 34.34 & 41.10 & 36.84 & 33.01 & 68.63 & 71.97 & 11.26 \\
\rowcolor{shadecolor}MCF~\cite{wang2023mcf} & 26.97 & 40.19 & 40.12 & 36.52 & 54.17 & 50.23 & 7.46\\\midrule
{\bf TI-ST} & \textcolor{ForestGreen}{37.69} & \textcolor{ForestGreen}{45.31} & \textcolor{ForestGreen}{50.93} & 40.87 & 66.56 & \textcolor{ForestGreen}{74.07} & \textcolor{ForestGreen}{16.18} \\ 
& \textcolor{gray}{\scriptsize{(+22.27)}} & \textcolor{gray}{\scriptsize{(+7.46)}} & \textcolor{gray}{\scriptsize{(+28.06)}} & \textcolor{gray}{\scriptsize{(+16.79)}} & \textcolor{gray}{\scriptsize{(+14.17)}} & \textcolor{gray}{\scriptsize{(+8.14)}}\\
\bottomrule
\end{tabular}

\end{table}


%% file: 05_conclusion.tex
\section{Conclusion}
\label{sec: conclusion}

We proposed a novel self-training framework with a self-assessment strategy for pseudo-label reliability, namely ``Transformation-Invariant Self-Training'' (TI-ST). This method uses transformation-invariant highly-confident predictions in the target dataset by considering an ensemble of high-confidence predictions from transformed versions of identical inputs. We experimentally show the effectiveness of our approach against numerous existing methods across three different source-to-target segmentation tasks, and when using different model architectures. Beyond this, we show that our approach is resilient to changes in the methods hyperparameter, making it well-suited for different applications. 